\RequirePackage{amsmath,amssymb}

\documentclass[runningheads]{llncs}

\input{config.sty}
\input{glossary.sty}

\begin{document}

\title{Abstract Hardware Grounding towards the Automated Design of Automation Systems}

%

\author{Yu-Zhe Shi \and
Qiao Xu \and
Fanxu Meng \and
Lecheng Ruan$^{\textrm{\Letter}}$ \and
Qining Wang$^{\textrm{\Letter}}$}
\authorrunning{Y.-Z. Shi et al.}
%
\institute{College of Engineering, Peking University, Beijing, 100871, China \\
$\textrm{\Letter}$ \email{\{ruanlecheng, qiningwang\}@pku.edu.cn}}

\maketitle              

\begin{abstract}
Crafting automation systems tailored for specific domains requires aligning the space of human experts' semantics with the space of robot executable actions, and scheduling the required resources and system layout accordingly. Regrettably, there are three major gaps, fine-grained domain-specific knowledge injection, heterogeneity between human knowledge and robot instructions, and diversity of users' preferences, resulting automation system design a case-by-case and labour-intensive effort, thus hindering the democratization of automation. We refer to this challenging alignment as the \emph{abstract hardware grounding} problem, where we firstly regard the procedural operations in humans' semantics space as the abstraction of hardware requirements, then we ground such abstractions to instantiated hardware devices, subject to constraints and preferences in the real world --- optimizing this problem is essentially standardizing and automating the design of automation systems. On this basis, we develop an automated design framework in a hybrid data-driven and principle-derived fashion. Results on designing self-driving laboratories for enhancing experiment-driven scientific discovery suggest our framework's potential to produce compact systems that fully satisfy domain-specific and user-customized requirements with no redundancy.

\end{abstract}

\section{Introduction}

Designing autonomous systems for expertise-intensive tasks, such as self-driving laboratories for scientific discovery~\cite{bedard2018reconfigurable,steiner2019organic,mehr2020universal,rohrbach2022digitization,burger2020mobile,szymanski2023autonomous}, dark factories for advanced manufacturing~\cite{buchner2023vision,kusiak2017smart,okwudire2021distributed}, and robot systems for clinical practices~\cite{jin2023injectable,sheetz2020trends,marcus2024ideal}, is essentially a non-trivial effort --- as robots need to act based on the knowledge of domain experts, there remains a substantial gap in translating human-generated domain-specific knowledge, \ie, semantics, into robot executable instructions, \ie, actions. In specific, there are three major obstacles. First, semantics is usually represented unstructurally, such as in \ac{nl} text, while actions require structural representation for precise parsing into step-by-step executable atomic units. Second, semantics is usually latent, with a large proportion of tacit knowledge only shared by domain experts, while actions require explicit representation for execution without further interpretation. Third, semantics is homogeneous across humans because they share the general actions functioned by the human body, while robots' actions are heterogeneous and tailored for specific functions. As a result, designing such automation systems requires a fine-grained integration between the highly diverged background knowledge of domain experts and automation experts~\cite{shi2023perslearn}, leading to a prohibitively expensive, case-by-case, and labor-intensive effort~\cite{shi2024autodsl}, which hinders the broader engineering practices of automation systems across various domains. 

Currently, in the design automation systems with high domain barriers, such as self-driving laboratories, experts make considerable efforts in bridging the gap between semantics and actions. They hand-craft pre-defined \acp{api} that provide a seamless one-to-one mapping from semantics to atomic actions, making the hardware system configuration aligned with domain-specific knowledge. For example, the Chemputer architecture for \ac{iot} based automated chemical synthesis and the KUKA robot platform for mobile manipulation represent two mainstream schools of thought~\cite{mehr2020universal,burger2020mobile}. In the former, researchers empirically tailor an \ac{iot} pipeline according to their focused scope through trial-and-error hacking and then directly pair the devices with semantics as \acp{api} for actions. In the latter, researchers use off-the-shelf commercial devices partially in the workflow and simply enjoy the \acp{api} developed by solution providers in system design. For both approaches, the cost of implementation is high but acceptable, as domain experts, such as chemists, build an automated chemical synthesis system only for their own use~\cite{christensen2021automation,el2023balancing,seifrid2022autonomous}. Consequently, this is a ``once and for all'' effort, and all related researchers can make fine-tunes to apply the system to a vast series of succeeding productions. 

Rather than the ``once and for all'' perspective of domain experts, from the position of the automation community, the gap toward the aligned semantics and action spaces substantially hinders the broader application of automation systems. In particular, this obstacle arises due to three major factors: (i) domain-specific semantic and action spaces; (ii) heterogeneity between semantic and action spaces; and (iii) specialized end-users' preferences. The three facets specify the requirements for system design.

\paragraph{Specific domains} 

Taking self-driving laboratory systems as an example, as Genetics and Bioengineering, two subdomains under Biology Sciences, share some commonalities with each other, yet the distribution of semantic actions for Genetics experiments significantly diverges from that for Bioengineering experiments, resulting in significant differences regarding the required hardware devices. Given the high specificity of domains, the target research scopes matter. Constrained by limited budget, each research group expects to get an automation system that is tailored compactly according to their target research scope, rather than applying general devices with functional redundancy~\cite{christensen2021automation}. For example, a system for Genetics experiments should not include any redundant features solely designed for Bioengineering.

\paragraph{Heterogeneous spaces} 

The mapping from the semantic space to the action space is not trivally one-to-one for general procedures and hardware devices. This issue usually arises when integrating unstandardized devices functioning as atomic operations into the pipeline, which greatly enlarges and refines the combinatorial space of functions compared to that of off-the-shelf devices~\cite{volk2024performance}. This effort brings heterogeneity into space mapping, resulting in three conditions: (i) One-to-many mapping, where a subgoal in semantic space may consist of several steps to be executed on different atomically functioned devices; (ii) Many-to-one mapping, where several subgoals in semantic space may be executed on the same device by several atomic functions; and (iii) Latent dependency, where several subgoals in semantic space are dependent on others and are thus omitted in procedures. These three conditions highlight the dismatch of granularities between semantic and action spaces.

\paragraph{User-centered preferences} 

The system is expected to be optimized in a user-centered manner and should be responsible for user preferences such as flexibility, cost, reliability, throughput, and efficiency. This results in a complicated system with a mixture of diverse hardware devices. For example, a decision is required on whether to implement a procedure by sequencing several atomic hardware devices in a fixed \ac{iot} pipeline, which is reliable but inflexible; or by employing a mobile robotic arm to transport reagents in containers to an off-the-shelf device, which is flexible yet costly~\cite{el2023balancing}. Another decision to make is whether to extend the system's parallelism capacity, which means redundant devices and communication channels, or to keep the system simple. Such decisions are made according to users' preferences and are intrinsically trade-offs on a Pareto frontier of optimizing multiple objectives.

Therefore, designing such an automated system requires a joint optimization across domain and automation knowledge, which are generally difficult due to the high cost of cross-domain collaboration. This highlights the significance of bridging the gap between human semantics and robot actions in a standardized automated way.

\begin{figure}[t]
    \centering
    \includegraphics[width=\linewidth]{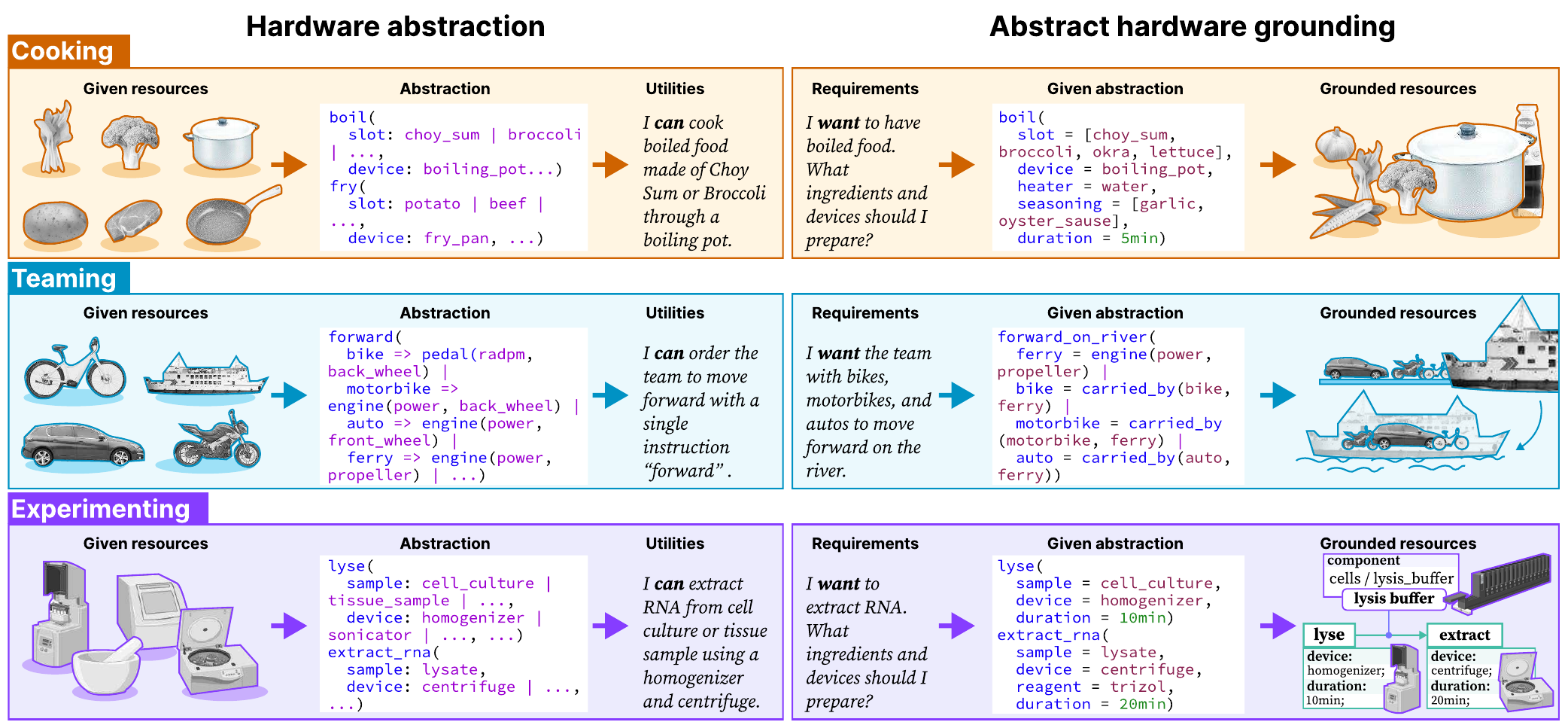}
    \caption{\textbf{Illustration of the paired inverse problems --- hardware abstraction and \ac{ahg}}}
    \vspace{-\baselineskip}
    \label{fig:motivation}
\end{figure}

Intuitively, this problem can be considered as an \emph{inverse} problem of extracting the hardware abstractions from given hardware specifications for building maximal common input-output interfaces that support control on abstract operations, which has been extensively studied in computer science~\cite{hennessy2011computer}. Indeed, the challenge is to ground given abstract operations for figuring out a minimal hardware configuration to execute the operations. We refer to this problem as the \ac{ahg} problem. Analogously, the \ac{ahg} problem is like tailoring a ``body'' for a robot that excels in a specific set of tasks, rather than developing available tasks for a given robot body. See \cref{fig:motivation} for illustration.

\begin{figure}[t]
    \centering
    \includegraphics[width=\linewidth]{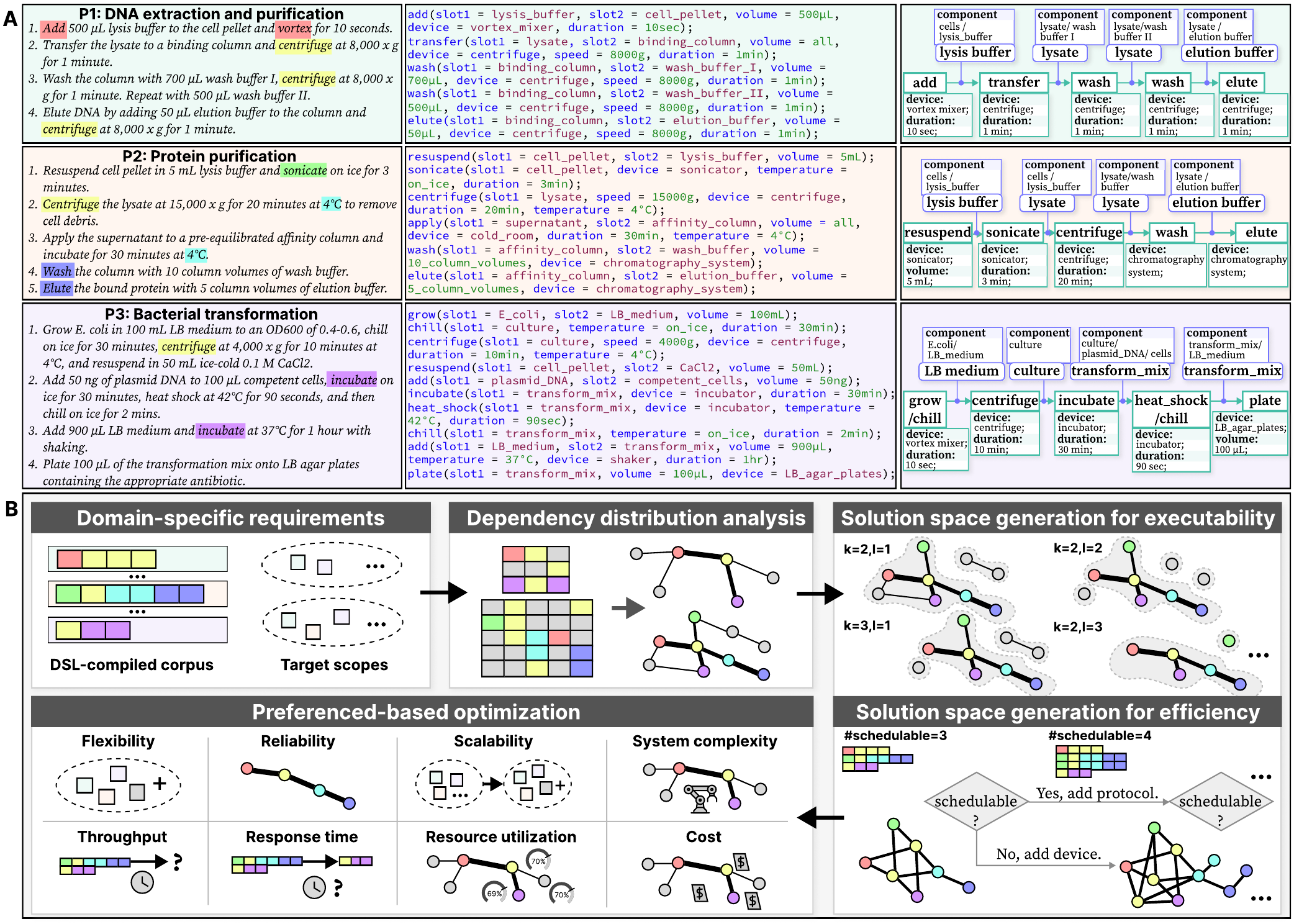}
    \caption{\textbf{Illustration of the \ac{ahg} problem.} \textbf{(A)} Demonstration of the original protocols and the protocols compiled into \ac{dsl} programs with corresponding protocol dependence graphs, serving as the hardware abstraction. \textbf{(B)} Workflow of the pipeline solving the \ac{ahg} problem.}
    \vspace{-\baselineskip}
    \label{fig:pipeline}
\end{figure}

In this work, we propose the initial proof-of-concept framework for solving the \ac{ahg} problem, which \textbf{aligns} the semantic and action spaces (see \cref{fig:pipeline}B), and \textbf{optimizes} automation systems' configurations according to specific domains, available devices, and multifaceted user preferences. First, \ac{nl}-described procedures within the scope of interest are compiled to \ac{dsl} programs for a precise and fine-grained structural knowledge representation. The target \acp{dsl} are customized efficiently according to domain-specific corpus through the AutoDSL tool developed by Shi \etal~\cite{shi2024autodsl}. Afterwards, the \ac{ahg} problem can be reduced to \emph{optimizing the hardware for executing the target scope of \ac{dsl} programs}, which recalls us with the lessons learned from computer science --- original procedures are user-centered high-level programs and the compiled \ac{dsl} programs are their corresponding \ac{isa} programs; the target automation system is a \ac{cpu} designed under specific requirements for executing the \ac{isa} programs. Interestingly, the current trend of replacing a general \ac{cpu} with an \ac{asic} tailored for users' requirements is in line with our customization of automation systems, which implies that the optimal design choice for the two efforts should be somewhat converged. In this context, our framework is inspired by automated \ac{asic} design on the high-level, applying both data-driven and theory-derived approaches. The framework adjusts the integration of \ac{iot} pipelines, \ie, pipelines in \acp{cpu}, and mobile robot transporters, \ie, bus lines in \acp{cpu}, driven by the distribution of dependency between actions, \ie, instructions in the \ac{isa}. It also balances the trade-offs on the multiple objectives according to users' preferences, derived from classic computer science problems such as resource scheduling and parallelism of programs. By reducing the empirical domain-specific design space to a computable design space, we can guarantee that the framework is grounded in established theories in computer architecture~\cite{aho1972theory,abelson1996structure,hopcroft2001introduction,bryant2011computer}, which sets up a solid foundation for the framework. 

We assess the practical utility of our framework regarding two levels: (i) the \textbf{executability} on the required set of tasks; (ii) the \textbf{efficiency} when executing tasks on a large scale. Specifically, we validate the framework on designing self-driving laboratories for experiment-driven scientific discovery. We find that the framework is capable of designing systems that maximize users' requirements subject to practical constraints. Meanwhile, the solutions are appropriately selected over trade-offs such as Flexibility \vs Reliability and Response time \vs System complexity, according to user preferences.

\section{The abstract hardware grounding problem} 

In this section, we introduce the \ac{ahg} problem. To concretize the concepts, we instantiate the problem to the task of designing self-driving laboratories and limit the problem scope into translating procedural knowledge on experiments to executable actions. 

\subsection{\ac{dsl} as hardware abstraction}

Procedural knowledge, \aka ``how-to knowledge'', is a type of knowledge describing the procedure to achieve an objective step-by-step. Experimental protocols, \ie, descriptions of the procedural knowledge on conducting scientific experiments, are highly constrained procedures to be executed in complex, resource-limited, and vulnerable laboratory environments. Those properties lead to the requirement of structural representation rather than \ac{nl} for hi-fidelity communication between the protocol author and the experiment executor~\cite{shi2024autodsl,baker20161,mcnutt2014reproducibility,munafo2017manifesto,freedman2015economics,baker2021five}. Thus, \acp{dsl} are leveraged to represent protocols as the most expressive form of structural knowledge representation~\cite{shi2024autodsl,fowler2010domain,mernik2005and}.

\ac{dsl} programs for protocol representation majorly come in the imperative programming paradigm~\cite{shi2024autodsl,steiner2019organic,mehr2020universal,wang2022ulsa,kearnes2021open,ananthanarayanan2010biocoder,strateos23autoprotocol}, where each unit of program is an atomic operation in the experiment workflow, with (i) preconditions indicating the input reagents; (ii) postconditions entailing the expected output products or observations; and (iii) parameters determining conditions of that operation, such as temperature, acidity, and duration. The operations encapsulate execution details into precondition and postcondition parameters, serving as the vehicle of hardware abstraction. Domain-specific knowledge is thus encoded as semantic constraints into the \acp{dsl}, such as the set of operations tailored for the domain with parameters only related to those operations. The relation between postconditions and preconditions from different atomic operations define the dependency between them --- if some products of \texttt{operation A} is included in the reagents of \texttt{operation B}, we say that ``\texttt{B} is dependent on \texttt{A}''~\cite{ferrante1987program}. Such topological order can be straightforward in sequential control flow and become complex in non-linear control flows such as loop and parallel, which is the core information to guarantee the executability of the protocol~\cite{baker2021five}. As such dependency is latent in \ac{nl} based text, exploiting structural knowledge representation, particularly the \acp{dsl}, as the representation of hardware abstraction, is a rational and imperative design choice (see \cref{fig:pipeline}A).

\subsection{\ac{ahg} problem definition}

We regard the \ac{ahg} problem as bridging the gap between the hardware abstraction on the semantics level and the target hardware system layout on the action level. 

\paragraph{Input}

Let $\mathcal{C}=\{\mathbf{c}_1,\mathbf{c}_2,\dots,\mathbf{c}_{|\mathcal{C}|}\}$ be the set of protocols of the target domain indicating the user's research scope. We denote each protocol compiled by \ac{dsl} as $\texttt{prog}(\mathbf{c})=\langle \texttt{op}_1,\texttt{op}_2,\dots,\texttt{op}_{|\texttt{prog}(\mathbf{c})|}\rangle\in\texttt{prog}(\mathcal{C})$. Let the operation set of the \ac{dsl} be $T_{\texttt{op}}$, for all $\texttt{op}_i,\texttt{op}_j\in T_{\texttt{op}}$ shown across protocols in the protocol set $\mathcal{C}$, we profile the distribution of the operation dependence relation by: 
\begin{equation}
    p(T_{op}|\mathcal{C})=\sum_{\mathbf{c}\in\mathcal{C}}\sum_{\texttt{op}_j\in T_{\texttt{op}}}\sum_{\texttt{op}_i\in T_{\texttt{op}}}p(\texttt{op}_i\prec\texttt{op}_j|\texttt{op}_i, \texttt{prog}(\mathbf{c})),
\end{equation}
where $\prec$ specifies the predecessor relation. To note, we leverage the conditional distribution $p(\texttt{op}_i\prec\texttt{op}_j|\texttt{op}_i)$ to reduce the bias introduced to $p(\texttt{op}_i\prec\texttt{op}_j)$ by $p(\texttt{op}_i)$.

\paragraph{Target Output}

The target design layout is represented as a directed graph with properties $\mathcal{L}=(\mathcal{A},\mathcal{R};\rho)$, where $\mathcal{A}=\{\dots a_i,\dots,a_j,\dots\}$ is the set of actions being grounded from the hardware abstraction layer, \ie, functional hardware devices, as vertices, $\mathcal{R}=\{\dots,r_{i,j},\dots\}$ is the set of connections between functional devices as directed edges, and $\rho(r_{i,j})$ is the property corresponding to the connection $r_{i,j}\in\{\texttt{grouped},\texttt{associated},\texttt{unconnected}\}$, with each property indicating the status quo of the connection between two devices: be connected directly with the same pipeline, \ie, \texttt{grouped}, be integrated indirectly with manipulation interfaces for mobile robotic arms, \ie, \texttt{associated}, or be simply unconnected, \ie, \texttt{unconnected}.

\subsection{Optimization on the grounded hardware layout} 

The final grounded hardware layout considers hard constraints by the target scope of research and user preferences, which can be denoted as $\mathcal{L}=(\mathcal{A},\mathcal{R};\rho|\mathcal{C},\mathcal{E},\mathcal{S})$, where $\mathcal{C}$ is the set of protocols required to be executed on the system, $\mathcal{E}(\cdot)$ is the combination of users' objective functions regarding executability, and $\mathcal{S}(\cdot)$ is the combination of users' objective functions regarding efficiency. To note, in general cases, all protocols in $\mathcal{C}$ work for the same domain such as Chemical Synthesis and can be translated to the single \ac{dsl} according to that domain. Thus, the overall objective function for optimizing the grounded hardware layout can be:
\begin{equation}
    \begin{split}
        \max\limits_{\mathcal{L}}\quad &\big(\mathcal{E}(\mathcal{L}),\mathcal{S}(\mathcal{L})\big)\\
        s.t.\;&\text{Executable(}\mathcal{C},\mathcal{L}\text{)},
    \end{split}
\end{equation}
where $\text{Executable(}\cdot,\cdot\text{)}:\mathcal{C}\times\mathcal{L}\mapsto \{\text{True}, \text{False}\}$ is a evaluation function on the executability of all protocols in $\mathcal{C}$ on the resulting hardware system. The executability verification is conducted by a virtual environment that traverses on the program dependence graphs of the corresponding \ac{dsl} programs~\cite{ferrante1987program,dourish1996freeflow}.

\section{Solving \ac{ahg} for executability}

In this section, we introduce the initialization and optimization for adapting the grounded hardware layout to users' requirements regarding executability. Then we validate the methodology through configuring the hardware system for conducting bench scale experiments in experimental sciences. We report quantitative results and discuss several preference-based trade-offs during the optimization through case studies.

\subsection{Initialization}

The elementary requirement for the hardware configuration is to at least support all the protocols in the scope of research. This establishes the foundation for making abstraction from the action level and for further optimization under user preferences. 

\begin{figure}[t]
    \centering
    \includegraphics[width=\linewidth]{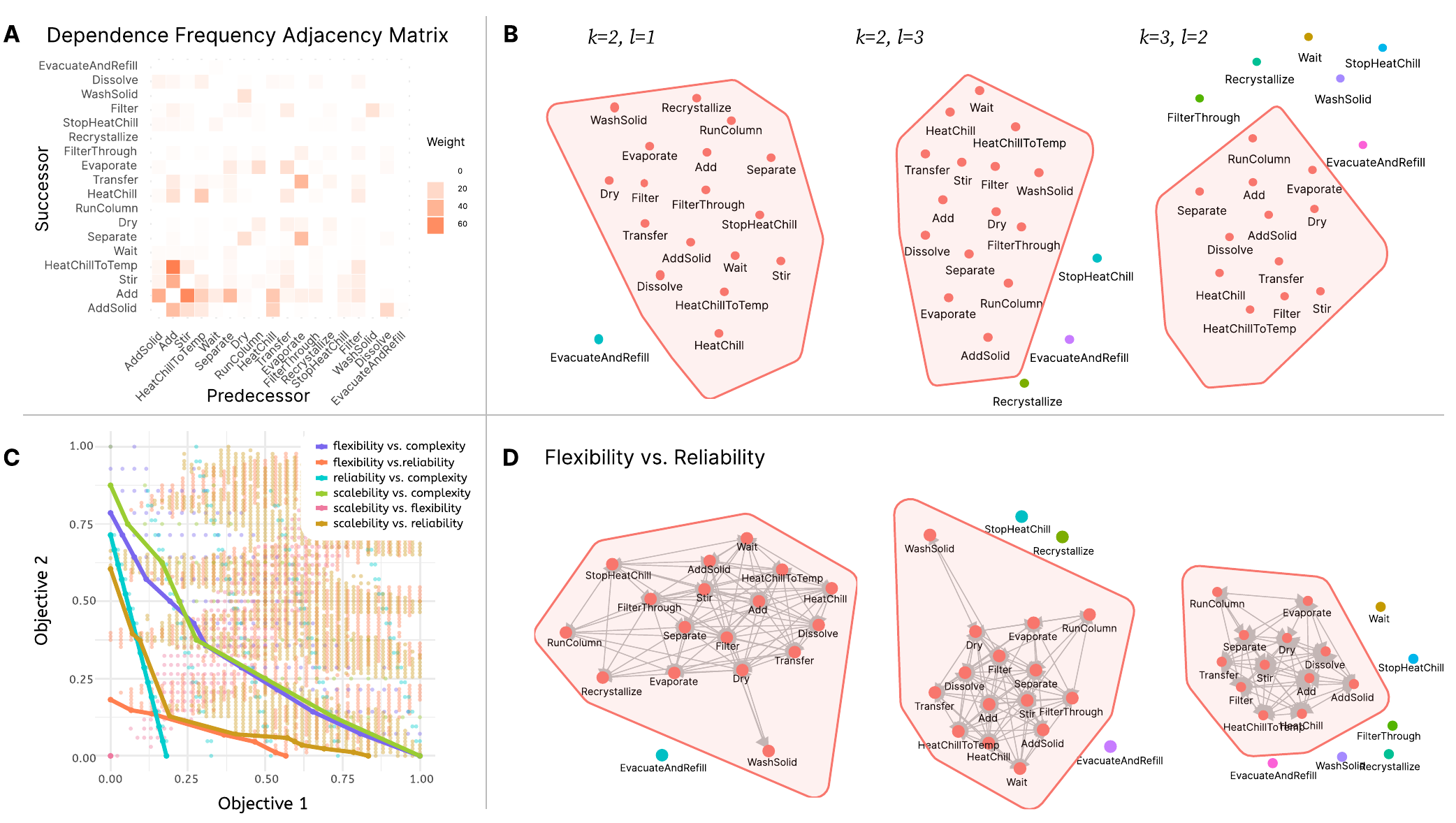}
    \caption{\textbf{Results on executability level.} \textbf{(A)} Distribution of operation dependency (adjacent matrix). \textbf{(B)} Showcases of k-minimal cut initialization. \textbf{(C)} Illustration of the Pareto frontiers in multi-objective optimization on executability level. (Two axis are a pair of normalized objectives) \textbf{(D)} Showcases of the resulting layouts by different trade-offs.}
    \vspace{-\baselineskip}
    \label{fig:res-executability}
\end{figure}

As the hardware abstraction encodes the distribution of the atomic operations, which is determined by the target set of protocols for execution, we employ a data-driven approach to obtain the distribution. First, we translate \ac{nl} based protocols into \ac{dsl} programs by the method used for AutoDSL utility assessment~\cite{shi2024autodsl}. Then, we figure out the workflow dependency graph of each protocol through postcondition-precondition verification, which essentially simulates each experiment step to check executability. Finally, we calculate the frequency of every ordered pair of operations (see \cref{fig:res-executability}A). To note, the execution order of a pair is asymmetric, \ie, \texttt{operation A -> operation B} ($a \prec b$) and \texttt{operation B -> operation A} ($b \prec a$) are two distinct pairs. 

Based on our intuition on pipelines and mobile robots, our feasible solution space is consist of system configurations with different proportions of pipeline-connected devices and robot-transportation-communicated devices. We assume that all devices are connected in pipelines by default. With the intuition that two set of actions without significant dependency should be separated for necessary flexibility, we reduce the problem of breaking one pipeline into multiple pipelines communicating by robot transportation as figuring out the $k$ minimum cuts on a weighted graph~\cite{goldschmidt1994polynomial}\footnote{The statistical significance of the dependency in one direction between a pair of actions is sufficient to put them in the same group without considering the other reversed direction. Based on this fact, we simplify the problem to calculate the $k$ minimum cuts on the undirected weighted graph, where the weight of the edge between $a_i$ and $a_j$ is $\max\{r_{i,j},r_{j,i}\}$. We only consider the directed weighted edges after determining the grouping of actions, to generate directed connections for the final layout.}. As the minimum-k cut algorithm divides the graph into $k$ connected components and recursively divides every component into $k$ components, there also exists $l$ possible hierarchies of recursion --- when $l$ is small, such as $l=1$, there is no recursion hierarchy and the actions are only divided into $k$ subsets; as $l$ increases till the maximum depth of recursion, the actions are splitted into smaller and smaller subsets. Consequently, $k$ and $l$ both affect the choice between pipelines and mobiles, spanning the full configuration space for preference-based optimization (see \cref{fig:res-executability}B).    

\subsection{Preference-based optimization}\label{subsec:opt}

As there are many feasible grounded hardware layouts given the target research scope, the exact system configuration should be specified according to user requirements. Here, we summarize five common factors that domain experts usually consider regarding self-driving laboratories as candidate objectives for optimization on executability~\cite{christensen2021automation,el2023balancing}: \emph{Flexibility}, \emph{Reliability}, \emph{Scalability}, \emph{System complexity}, and \emph{Cost}. To note, some of the objectives may conflict with each other.

\paragraph{Flexibility}

The maximal set of protocols $\mathcal{C}^*\supset\mathcal{C}$ that can be exactly executed by the system without any modifications, in addition to the required set of protocols $\mathcal{C}$. Usually higher is preferred. Thus we have: 
\begin{equation}
    \begin{split}
        \max\limits_{\mathcal{L}}\quad & |\mathcal{C}^*\setminus\mathcal{C}|\\
        s.t.\;&\text{Executable(}\mathcal{C}^*,\mathcal{L}\text{)}.
    \end{split}
\end{equation}

\paragraph{Reliability}

The maximal proportions of operations in the target scope of protocols that can be executed in static pipelines rather than by dynamic robot transportation. Usually higher is preferred. Thus we have:
\begin{equation}
    \begin{split}
        \max\limits_{\mathcal{L}}\quad & |\{r|r=\texttt{grouped},r\in\mathcal{R}\}|\\
        s.t.\;&\text{Executable(}\mathcal{C},\mathcal{L}\text{)}.
    \end{split}
\end{equation}

\paragraph{Scalability}

The minimal modification to be made when extending the system from executing the current target scope of protocols $\mathcal{C}$ to executing a broader set of protocols $\mathcal{C}^{**}\supset\mathcal{C}$, which is evaluated by the edit distance. Usually lower is preferred. We denote the modified layout as $\mathcal{L}'=\texttt{mdfy}(\mathcal{L})$, where $\texttt{mdfy}(\cdot): \mathcal{L}\mapsto\mathcal{L}$ is a \ac{fsm} based function manipulating the vertices and edges of the graphs with three operators: \texttt{insertion}, \texttt{deletion}, and \texttt{substitution}. Thus we have: 
\begin{equation}
    \begin{split}
        \min\limits_{\mathcal{L}}\quad & \text{dist}(\mathcal{L}', \mathcal{L})\\
        s.t.\;&\text{Executable(}\mathcal{C}^{**},\mathcal{L}'\text{)}.
    \end{split}
\end{equation}

\paragraph{System complexity}

Sum of the total number of nodes, the total number of pipeline edges, and the total number of robot edges multiplied by the \dof of each robot respectively according to the resulting system layout graph. Usually lower is preferred. The \dof of the system layout with mobile robot(s) is calculated as $\text{DoF}(r') = c(r') + n(r') + \gamma(r')$, where $c(r')$ is the number of \texttt{associated} connections regarding the robot(s), $n(r')$ indicates the moving range of the robot(s), $\gamma(r')$ is the complexity coefficient according to the involved robot(s). Thus we have:
\begin{equation}
    \begin{split}
        \min\limits_{\mathcal{L}}\quad & |\mathcal{A}|+|\{r|r=\texttt{grouped},r\in\mathcal{R}\}|+\sum_{r'=\texttt{associated},r'\in\mathcal{R}}\text{DoF}(r')\\
        s.t.\;&\text{Executable(}\mathcal{C},\mathcal{L}\text{)}.
    \end{split}
\end{equation}

\paragraph{Cost}

Sum of the total price of hardware devices, the total price of mobile robots, and the total price of pipelines according to the resulting system layout, denoted as $\text{Cost(}\mathcal{L}\text{)}$. Usually lower is preferred. The prices of devices are extracted from a published general laboratory guideline for experimental sciences~\cite{mcmahon2008analytical}. 

\subsection{Results}\label{subsec:res-executability}

\paragraph{Materials}

We consider the protocols of real experiments, which are retrieved from Protocolexchange published by Nature\footnote{\url{https://protocolexchange.researchsquare.com/}}. There are 171 protocols in the target research scope $\mathcal{C}$ ($33\%$ of all protocols). The sizes of maximal protocol set $\mathcal{C}^*$ range from 461 ($89.5\%$) to 495 ($98\%$). The total number of protocols for scaling $\mathcal{C}^{**}$ is 515.

\paragraph{Multi-objective optimization}

To investigate the relations among the five candidate objectives, we construct a multi-objective optimization function by combining all objectives and constraints, resulting the Pareto frontier in \cref{fig:res-executability}C. We consider the dyadic relations between pairs of objectives. We find that there are trade-offs between \emph{Flexibility-Reliability} and \emph{Scalability-Reliability}. Also, \emph{System complexity} trade-offs with \emph{Flexibility}, \emph{Reliability}, and \emph{Scalability}.

\paragraph{Discussion}

The \emph{Flexibility-Reliability} trade-off comes from the choice between mobile robots and pipelines, where the former introduces higher error rates compared to the latter. Similarly, the \emph{Scalability-Reliability} trade-off raises from the consideration of adaptable devices, which lead to the isolation of frequently-used devices from the pipelines, thus introducing additional \dof and reducing system reliability. In comparison with \emph{Reliability}, \emph{Flexibility} and \emph{Scalability} are compatible objectives. Moreover, \emph{System complexity} possesses non-monotonic relations among \emph{Flexibility}, \emph{Reliability}, and \emph{Scalability}, as alternating between mobile robots and pipelines is not a simple one-to-one mapping --- removing a mobile robot with several interactive interfaces from the system may cause exponential expansion of pipelines (see \cref{fig:res-executability}D). 

\section{Solving \ac{ahg} optimization for efficiency}

In this section, we introduce the initialization and optimization for adapting the grounded hardware layout to users' requirements regarding efficiency. Then we validate the methodology through designing the layout for conducting pilot scale experiments in experimental sciences. We report quantitative results and discuss several preference-based trade-offs during the optimization through case studies.

\subsection{Initialization}

\begin{figure}[t]
    \centering
    \includegraphics[width=\linewidth]{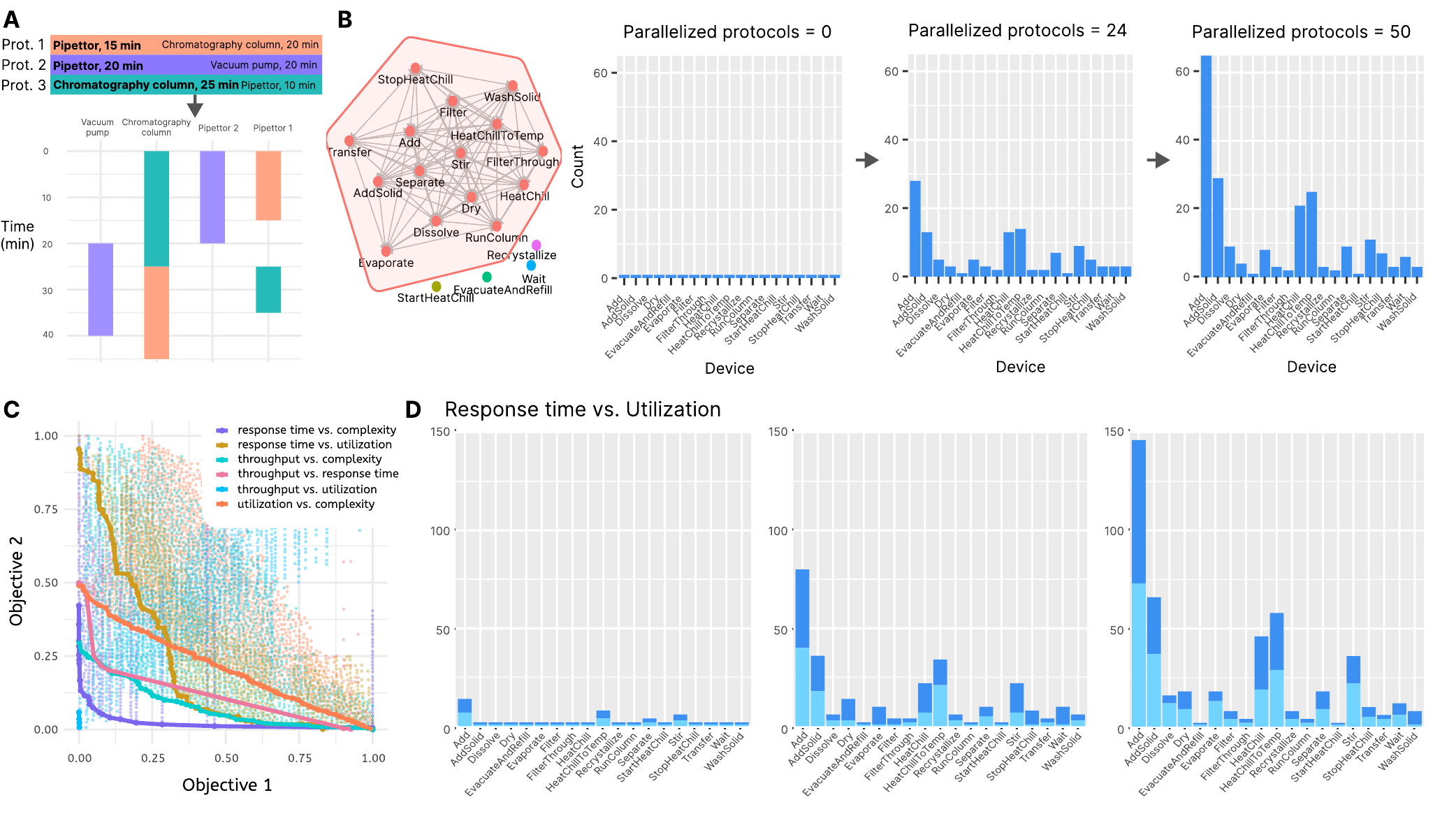}
    \caption{\textbf{Results on efficiency level.} \textbf{(A)} Intermediate process of resource scheduling. \textbf{(B)} Showcases of iterative algorithm initialization. \textbf{Left:} Layout of the starting point of the iterative algorithm. \textbf{Right:} Histograms of the required types and quantities of devices. \textbf{(C)} Illustration of the Pareto frontiers in multi-objective optimization on efficiency level. \textbf{(D)} Showcases of the resulting layouts by different trade-offs in histograms of the required types and quantities of devices. Color difference indicates the observation point switching from all \textbf{(dark)} to individual protocols.}
     \vspace{-\baselineskip}
    \label{fig:res-efficiency}
\end{figure}

Based on the guarantee of executability of a given set of protocols, \ie, the basic hardware abstraction, we consider the scaling-up of tasks, which requires scheduling and parallelization of execution. The key intuition is that, if we aim to apply the schedule the resources, we should at least have the hardware configuration to sufficiently support the possible scheduling strategies, just like \emph{``if you want to conduct scheduling on a computer cluster for high performance computing, you should at least have a multiple-node cluster, or all queues would be executed sequentially''} (see \cref{fig:res-efficiency}A). This serves as a starting point to consider resource scheduling and prune the over-redundancy system. 

We design an iterative algorithm to figure out the target hardware configuration. First, we obtain minimal layout that is able to execute all of the protocols in the given set without any additional objectives for optimization. Currently, the resulting system is capable to execute all of the protocols one-by-one without any considerations regarding efficiency. Then, we constrain the problem one step further. We randomly pick a pair of protocols from the set and employ a mainstream scheduling strategy, \eg, \emph{first-come-first-serve}, to schedule resources for executing this pair of protocols simultaneously on the current system. This step yields a three-fold choice: (i) if the scheduling strategy succeeds, which means that the current system configuration is sufficient to support scheduling, then we randomly pick another protocol into the current picked protocols, and return to the choice; (ii) if the scheduling strategy fails, which means that the current system configuration is insufficient to support scheduling, then we make minimal modification to the system by appending necessary resources, and return to the choice; (iii) if all protocols in the target set of protocols have been picked, cease the iteration and the resulting hardware layout is our target system configuration (see \cref{fig:res-efficiency}B).

The feasible solution space of the iterative algorithm is spanned by the extend of scheduling applying to the system. As aforementioned, the extreme condition is that all of the target protocols are expected to be scheduled on the system. Intuitively, we can relax this condition to one that scheduling only a part of the target protocols $\mathcal{C}^S\subset\mathcal{C}$ on the system. The proportion of protocols in the target protocol set to be scheduled yields a spectrum of systems with different capacities of supporting scheduling. To note, the selection of protocols from the first iteration affects the entire iterative process. To maintain the randomness of the algorithm, we sample several \emph{representative} protocols $\mathbf{C}^S$ are the initialization respectively and run repeated random trials accordingly. The representative protocols are sampled around the centroids of the clusters generated by dimension reduction from the embedding space of the protocols~\cite{beltagy2019scibert}. Consequently, $|\mathbf{C}^S|/|\mathcal{C}|$ spans the full configuration space for preference-based optimization.

\subsection{Preference-based optimization}

We summarize five common factors that domain experts usually consider regarding self-driving laboratories as candidate objectives for the optimization on executability~\cite{el2023balancing}: Throughput, Response time, Resource utilization, System complexity, and Cost. To note, some of the objectives may conflict with each other. We assign the unspecified execution duration of operations with queries in a published laboratory manual~\cite{national2011prudent}.

The scheduling is conducted over the same operation instances with the same precondition and postcondition across all of the protocols in $\mathcal{C}$, denoted as $\texttt{op}_i\texttt{[s,e]}$. Accordingly, we define a ready queue $Q(\texttt{op}_i\texttt{[s,e]})$ for operations waiting for available execution resources and an active queue $Q'(\texttt{op}_i\texttt{[s,e]})$ for operations in execution.

\paragraph{Throughput}

The expected volume of work, \ie, the expected number of operations, that can be pushed into all ready queues and popped from all active queues over a given time interval $\Delta T$. Usually higher is preferred. Thus we have:
\begin{equation}
    \begin{split}
        \max\limits_{\mathcal{L}}\quad & \sum_{\texttt{op}_i\texttt{[s,e]}\in\mathcal{C}} \big|\Delta_{\texttt{push}}Q(\texttt{op}_i\texttt{[s,e]})[t]\big| + \big|\Delta_{\texttt{pop}}Q'(\texttt{op}_i\texttt{[s,e]})[t]\big|\\
        s.t.\;&\text{Executable(}\mathcal{C},\mathcal{L}\text{)},\;\text{Cost(}\mathcal{L}\text{)}\leq C_M.
    \end{split}
\end{equation}

\paragraph{Response time}

The expected time consumed for an operation from being pushed into a ready queue to being popped out of an execution queue. Usually lower is preferred. Let $\text{TS(}\cdot\text{)}$ be the timestamp of an event. Thus we have:
\begin{equation}
    \begin{split}
        \min\limits_{\mathcal{L}}\quad & \sum_{\texttt{op}_i\texttt{[s,e]}\in\mathcal{C}} \text{TS(}Q'(\texttt{op}_i\texttt{[s,e]}).\texttt{pop}\text{)}-\text{TS(}Q(\texttt{op}_i\texttt{[s,e]}).\texttt{push}\text{)}\\
        s.t.\;&\text{Executable(}\mathcal{C},\mathcal{L}\text{)},\;\text{Cost(}\mathcal{L}\text{)}\leq C_M.
    \end{split}
\end{equation}

\paragraph{Resource utilization}

The expected occupation rate of each device in the system when completing protocols of the target protocol set over a given $\Delta T$. Thus we have:
\begin{equation}
    \begin{split}
        \max\limits_{\mathcal{L}}\quad & \frac{\big|\{Q'(\texttt{op}_i\texttt{[s,e]})|Q'(\texttt{op}_i\texttt{[s,e]})=\phi,\texttt{op}_i\texttt{[s,e]}\in\mathcal{C}\}\big|}{\big|\{Q'(\texttt{op}_i\texttt{[s,e]})|\texttt{op}_i\texttt{[s,e]}\in\mathcal{C}\}\big|}\\
        s.t.\;&\text{Executable(}\mathcal{C},\mathcal{L}\text{)},\;\text{Cost(}\mathcal{L}\text{)}\leq C_M.
    \end{split}
\end{equation}

\paragraph{System complexity}

The same as system complexity described in \cref{subsec:opt}.

\paragraph{Cost}

The same as cost described in \cref{subsec:opt}. 

\subsection{Results}

\paragraph{Materials}

We assign the full protocol set (515 protocols) to the target research scope $\mathcal{C}$. The maximum number of parallelized protocols is 50. We empirically set $\Delta T=5\times 10^4 s$ for \emph{Throughput}, $\Delta T=10^4 s$ for \emph{Resource utilization}, and the cost constraint $C_M=$ for \emph{Throughput}, \emph{Response time}, and \emph{Resource utilization}.

\paragraph{Multi-objective optimization}

To investigate the relations among the five candidate objectives, we construct a multi-objective optimization function by combining all objectives and constraints, resulting the Pareto frontier in \cref{fig:res-efficiency}C. We consider the relation of ten pairs of objectives. We find that there are major trade-offs between each two among \emph{Throughput}, \emph{Response time}, and \emph{Resource utilization}. Also, \emph{System complexity} trade-offs with \emph{Flexibility}, \emph{Reliability}, and \emph{Scalability}.

\paragraph{Discussions}

The trade-off between \emph{Throughput} and \emph{Response time} comes from the intrinsic conflict between global arrangement and local priority. When multiple execution sequences are scheduled tightly to maintain completeness of the target protocol set, the efficiency for executing one specific protocol is sacrificed accordingly. \emph{Throughput} and \emph{Resource utilization} conflict with each other because the former maintains redundancy for parallel operations, always causing device idle that is undesirable for the latter. The trade-off between \emph{Response time} and \emph{Resource utilization} is also raised by the demand of reserving available devices for specific experiments in advance, thus resulting in a population of idle devices. In line with \cref{subsec:res-executability}, \emph{System complexity} is non-monotonic among \emph{Throughput}, \emph{Response time}, and \emph{Resource utilization}, as appending duplicated devices or removing redundant devices are not simple one-to-one mappings, which usually leads to significant changes on connections of the resulting layout, \ie, alternations of pipeline connectivity or \dof of mobile robots (see \cref{fig:res-efficiency}D).

\section{General discussion}

In this work, we propose the initial proof-of-concept framework to solve the \ac{ahg} problem. Through our real-world experiments for designing domain-specific self-driving laboratories, we find that our framework is able to design automation systems that compactly satisfies specific domains, available devices, and multifaceted user preferences. Moreover, we test several trade-offs on the levels regarding executability and efficiency respectively, evaluating them with bench-scale and pilot-scale experimental requirements. We discuss the corresponding design principles discovered from the Pareto frontier of the multi-objective optimization. These results suggest that our framework has the potential to specialize automation systems for different domains broadly. 

\paragraph{On heterogeneity of automation systems}
Currently there is a lively discussion on whether \emph{humanoid robot}, the robot with a general embodied form to execute as diverse actions as humans can, is the ultimate answer of automation systems~\cite{fukuda2017humanoid}. We hold the position that specialized robots aimed for domain-specific tasks are irreplaceable and maybe essential for the embodiment of intelligence. While a humanoid robot may bring in blue-sky thinking of \ac{agi}, it must sacrifice performance and effectiveness in each specific task compared with the corresponding specified robot. From a macro perspective, given \textbf{limited resources}, there exists a Pareto frontier regarding multiple design requirements. If we expect the resulting robot excels at a domain-specific task and even outperforms human, the design should maximize the target requirement and minimize all other requirements. Such transition helps the robot change from ``is capable of diverse tasks but excels at none of them'' to ``is excelling at only one task''. This should be the natural, rational, and economic choice for engineering practice~\cite{shoval2012evolutionary}, where maximizing the overall utility is the pivot requirement. The proposed method in this paper gives an example of how the group optima is obtained by the teaming of heterogeneous robot ``experts''.

\paragraph{Yet another ``general'' solution for engineering practice}

Domain-specification inherently introduces cost. This shapes a dilemma between specificity and generality. On one side, automation systems tailored for specific domains excel at the corresponding tasks, while crafting such systems are case-by-case labour intensive efforts because fine-grained domain knowledge injection is not reusable, which makes the design process prohibitively expensive. On the other side, automation systems maintaining domain generality can be extremely complex and may loss performance on specific tasks due to the sacrifice on adaption of other tasks. To alleviate such dilemma, a practical way is to try to enhance the efficiency for tailoring domain-specific automation systems, such as domain knowledge injection, by both bottom-up data-driven approaches and top-down principle-derived methods. This idea motivates the development of the AutoDSL tool~\cite{shi2024autodsl}, and also the automated \ac{ahg} solving framework proposed in this paper. By integrating these two approaches, we may expect a fully automatic design process of domain-specific automation systems in the near future. The design process is an explicit two-stage framework, where the AutoDSL tool automatically designs the corresponding \ac{dsl} as hardware abstractions, and then the \ac{ahg} solver automatically designs the grounded hardware layout according to requirements. We hope these efforts would make domain-specification much more efficient and affordable, thus achieving the \emph{general} application of automation in a roundabout way.

\begin{credits}
\subsubsection{\ackname} This work was supported by the National Natural Science Foundation of China under Grant 91948302. The authors appreciate the support from Prof. Huamin Qu.

\subsubsection{\discintname}
The authors have no competing interests to declare that are relevant to the content of this article.
\end{credits}

\bibliographystyle{splncs04}
\bibliography{references}

\end{document}